\documentclass{elsevier-book}

\usepackage{graphicx,url}
\usepackage{array,booktabs,threeparttable,longtable}
\usepackage{cite}
\usepackage{enumitem}
\usepackage{algorithm,algorithmic}
\usepackage{lipsum}
\usepackage{titlesec}
\usepackage{xcolor}
\usepackage{adjustbox}
\usepackage{multirow}
\usepackage{float}
\usepackage{fancyvrb}
\usepackage{amsmath,amssymb,amsthm}
\usepackage[english]{babel}

\usepackage{listings}
\definecolor{bg}{rgb}{1,1,1}
\lstdefinelanguage{JavaScript}{
  keywords={break, case, catch, continue, debugger, default, delete, do, else, false, finally, for, function, if, in, instanceof, new, null, return, switch, this, throw, true, try, typeof, var, void, while, with},
  morecomment=[l]{//},
  morecomment=[s]{/*}{*/},
  morestring=[b]',
  morestring=[b]",
  sensitive=true
}
\theoremstyle{remark}

\begin{document}
\chapter{Incentive-Based Federated Learning: Architectural Elements and Future Directions}

\chapterauthors{Chanuka A.S. Hewa Kaluannakkage and Rajkumar Buyya}{%
Quantum Cloud and Distributed Systems (qCLOUDS) Lab \\
School of Computing and Information Systems \\
The University of Melbourne, Australia}

\begin{abstract}

Federated learning promises to revolutionize machine learning by enabling collaborative model training without compromising data privacy. However, practical adaptability can be limited by critical factors, such as the participation dilemma. Participating entities are often unwilling to contribute to a learning system unless they receive some benefits, or they may pretend to participate and free-ride on others. This chapter identifies the fundamental challenges in designing incentive mechanisms for federated learning systems. It examines how foundational concepts from economics and game theory can be applied to federated learning, alongside technology-driven solutions such as blockchain and deep reinforcement learning. This work presents a comprehensive taxonomy that thoroughly covers both centralized and decentralized architectures based on the aforementioned theoretical concepts.
Furthermore, the concepts described are presented from an application perspective, covering emerging industrial applications, including healthcare, smart infrastructure, vehicular networks, and blockchain-based decentralized systems. Through this exploration, this chapter demonstrates that well-designed incentive mechanisms are not merely optional features but essential components for the practical success of federated learning. This analysis reveals both the promising solutions that have emerged and the significant challenges that remain in building truly sustainable, fair, and robust federated learning ecosystems.

\end{abstract}



\newpage
\section{Introduction}

Federated learning (FL) has emerged as a revolutionary approach in machine learning (ML), enabling models to be trained collaboratively without sharing sensitive information with a central server. However, despite its advantages, federated learning suffers from an inherent problem that could adversely affect its widespread adoption, referred to as the participation dilemma. This chapter addresses the central issue of individual participation in federated learning systems, theoretically and empirically exploring incentive mechanisms designed to address this issue.

\subsection{The fundamental Participation Dilemma in Federated Learning}

Unlike conventional learning approaches, where data must be transferred to a central location for computation, federated learning allows data owners to collaboratively train their local machine learning models while keeping the private data on local devices without exposing it, thus preserving data confidentiality \cite{FL-challenges-and-future-direction, advances-and-open-problems-in-fl, survey-IM}. The concept was originally introduced to the public by Google as a means of improving mobile keyboard predictions without compromising user privacy \cite{pmlr-v54-mcmahan17a}. This emerging concept has been applied in numerous diverse domains, including healthcare, finance, telecommunications, and smart cities \cite{FL-challenges-and-future-direction}. 

The practical application of both Centralized and Decentralized Federated Learning (CFL and DFL) is inherently impeded by a fundamental issue in the participation dilemma, which threatens its viability \cite{IM-for-FL}. This participation dilemma is associated with several key issues. First, participating entities incur expenses for computational power and communication resources while facing potential privacy risks, even with the protocol that preserves privacy \cite{pmlr-v119-sim20a}. Second, there is no transparency regarding the total reward for the global model, nor is it shared equally among participants based on their contributions, which is pertinent in the context of CFL. The transparency issue becomes complex in the context of DFL since there is no central authority \cite{game-theory-based-BC}. Third, the benefits of participation are often delayed, uncertain, and shared among all participating entities. Fourth, participating entities may not have concrete trust in the system or in other participants. This cost-benefit imbalance creates a controversial situation when participation is voluntary \cite{survey-IM, pmlr-v130-fraboni21a, lin2019freeridersfederatedlearningattacks}.

The heterogeneity of participants in the FL system further escalates the problem. Data owners from resource-rich organizations, such as hospitals with high-quality datasets, may not wish to contribute, as this would involve sacrificing their computational power, energy, bandwidth, and, most importantly, their private data for the well-being of the remaining participants who contribute low-quality or irrelevant data. This describes the conflict between individual self-interest and the collective good. 

\subsection{Overview of Free-Rider Problem in FL Systems}

Collaborative parties in a distributed system should utilize their computational and communication power while adhering to a specific protocol defined by the system. Participating entities in a federated learning system contribute their local data, computational power, and communication resources. These participants can be categorized as follows: (i) \textit{obedient users} who strictly follow the system protocols without focusing on their own benefits, and (ii) \textit{rational users} who make decisions aimed at maximizing their own benefits. They may act outside system protocols if doing so is advantageous to them. \cite{overcomingfree-ridingbehaviorinpeer-to-peersystems}. However, obedient behavior is considered an unrealistic phenomenon in real-world distributed systems. 

In the context of federated learning, participating entities whose ultimate goal is to obtain a well-trained global or consensus model without contributing any data are referred to as \textit{free riders}. They benefit from the resources, public goods, or services of a collaborative system but do not wish to contribute their own resources or private data. Free riders manipulate model updates in ways that make it difficult for parameter servers in centralized FL settings, or peers in decentralized FL settings, to detect their behavior using a validation dataset. Several techniques have been explored to fool the system, including sending back the same model received by a free-rider (called plain free-riding), adding a small perturbation to the received model (called disguised free-riding), and subtracting two consecutive models to compute the gradients for a given round \cite{lin2019freeridersfederatedlearningattacks, pmlr-v130-fraboni21a}. Most of the time, the FL system converges, even in the presence of free riders. However, their participation still degrades the quality of the global model and creates unfairness for legitimate participants.

\subsection{Economic Rationale for Incentive Mechanisms}

In general terms, incentivization refers to encouraging an entity to take a specific action. Economic principles have long been used to regulate cooperation and resource sharing in distributed computing environments such as Grid computing and peer-to-peer (P2P) systems. In these settings, participants often act rationally and selfishly, requiring mechanisms that align individual interests with overall system efficiency. Earlier works introduced the notion of a computational economy, where market-based resource allocation, auctions, and credit systems were employed to manage CPU cycles, storage, and network bandwidth across heterogeneous administrative domains \cite{economicsmodelsbybuyya, grideconomy_buyya, buyya2002economicbaseddistributedresourcemanagement}. These economic approaches established key principles such as fairness, efficiency, and individual rationality criteria, which remain central to modern incentive design.

Building on these foundations, federated learning (FL) inherits similar challenges, as it relies on the voluntary participation of distributed clients with varying data quality and resource capacities. Accordingly, incentive mechanisms in FL extend these economic ideas to encourage sustainable participation and high-quality contributions, often by rewarding data value, computational effort, or reliability. The economic theory behind incentive mechanisms in federated learning is derived from game theory and mechanism theory \cite{overcomingfree-ridingbehaviorinpeer-to-peersystems}. These frameworks are used not only to compensate participants but also to encourage rational participants to contribute honestly and consistently to the FL system that aligns with their own interests.

Unless provided with proper incentives. Participating entities may behave as free-riders, especially in the absence of an incentive mechanism. Incentive mechanisms in federated learning become a critical aspect for the reasons mentioned earlier, which combine insights from economics, game theory, mechanism design, and distributed systems. Therefore, the goal is to develop mechanisms that fairly reward individual participants and the federated learning system while maintaining privacy and efficiency. It allows sustainable participation from data owners, eliminates the free-rider effect, and makes FL systems more attractive in the first place.

The motivation behind the need for a proper incentive mechanism is presented in Table \ref{tab:motivational_factors}.









\begin{table*}[!ht]
\centering
\begin{threeparttable}
\caption{Key Motivational Factors}
\label{tab:motivational_factors}
\resizebox{\textwidth}{!}{%
\begin{tabular}{@{}>{\centering\arraybackslash}p{0.2\textwidth}>{\centering\arraybackslash}p{0.35\textwidth}>{\centering\arraybackslash}p{0.4\textwidth}@{}}
\toprule
\textbf{Challenge} & \textbf{Description} & \textbf{Impact}  \\
\midrule
Voluntary Participation & Clients join FL networks voluntarily & Without incentives, participation rates drop significantly \\
Resource Consumption & Training requires computational power and bandwidth & Participants incur real costs without guaranteed benefits \\
Data Privacy Concerns & Clients worry about information leakage & Need assurance that participation is worthwhile \\
Quality Assurance & Ensuring high-quality contributions & Poor incentives lead to low-effort participation \\
Fairness & Fair evaluation and distribution among participants & Fairness ensures sustainable participation across the system \\
\bottomrule
\end{tabular}%
}
\end{threeparttable}
\end{table*}

Incentive mechanisms consist of two main problems. The first challenge is figuring out how to encourage and retain participants in a federated learning system from their perspectives. This mainly involves offering attractive rewards while still protecting their data privacy. The second challenge arises from the viewpoint of the federated learning task publisher, which requires evaluating each participant's contribution. As a result, designing an effective incentive mechanism is challenging because it must manage system fairness, encourage active participation, and ensure the long-term sustainability of the system.

\subsection{Chapter Roadmap and Contribution}

The rest of the chapter is organized as follows: It begins by examining the architectures of incentive mechanisms, including those for CFL and DFL. It then presents a systematic taxonomy of existing paradigms, including economic and game-theoretic approaches, as well as technology-based approaches, providing detailed coverage of incentive-oriented systems and applications. Finally, conclude with a discussion on existing challenges and future directions.


\section{Incentive-Based Federated Learning (IBFL) Architecture}

IBFL architecture must embed economic and game-theoretic strategies into the FL infrastructure to address the problems identified in the introduction section and to ensure sustainable participation throughout the learning process. These additional layers can extend the conventional FL process to form an Incentive-Based Federated Learning (IBFL) approach: (1) \textit{a contribution evaluation layer} that quantifies the quality and quantity of each participant's input, (2) \textit{an incentive calculation layer} that translates contributions into economic rewards, (3) \textit{a payment distribution layer} that allocates rewards earned by participants according to a fairness criterion, and (4) \textit{a trust and reputation tracking layer} that records historical reputations and trust earned by participating entities for future interactions. These components should operate seamlessly within the iterative training process without compromising the fundamental privacy assurances that make the federated learning paradigm attractive in the first place.

\subsection{IBFL Architecture for Centralized Federated Learning (C-IBFL)}

\begin{figure}[ht] 
    \centering
    \includegraphics[width=0.8\textwidth]{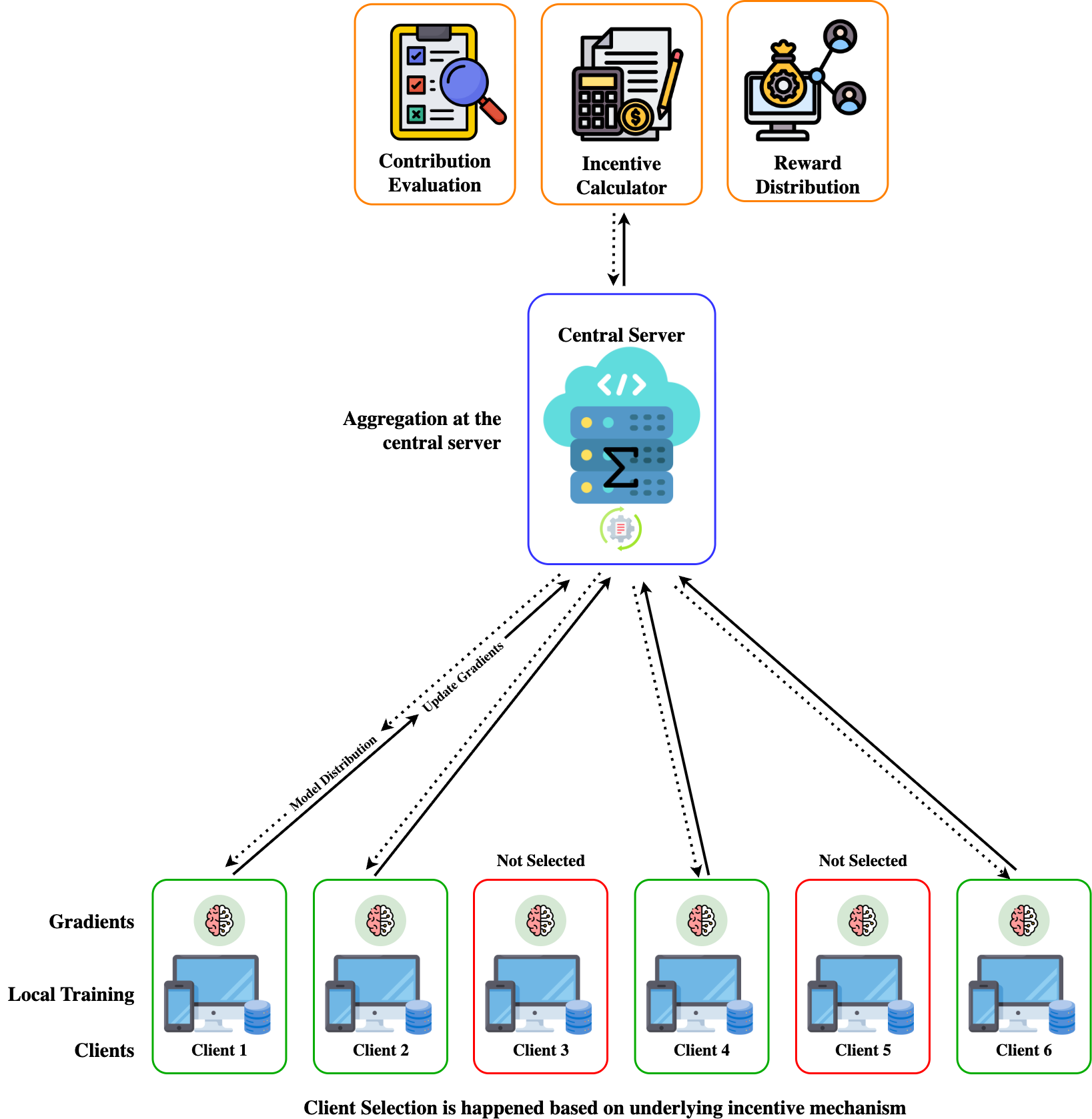} 
    \caption{C-IBFL architecture. Each client is selected to optimize the collective welfare of the system. Client selection happens according to the underlying incentive mechanism. The selected clients perform local training and share their gradients with the server. The server aggregates these gradients while simultaneously evaluating them and calculating the incentives, which are then distributed among the clients.}
    \label{fig:CIBFL}
\end{figure}

The Centralized Federated Learning with Incentives (C-IBFL) architecture maintains the classical client-server architecture. Unlike the vanilla FL, the server is also responsible for incentive management for all participating entities \cite{survey-IM}. Figure \ref{fig:CIBFL} represents the centralized federated learning architecture with integrated incentive principles. In this chapter, the term "clients" may be utilized as a synonym for participants.

\textbf{Central Server Architecture:} The server consists of four integrated subsystems. The first, the Model Aggregator, carries out the core FL function by combining clients’ model updates using an aggregation algorithm defined by the system, such as FedAvg \cite{FedAvg} or weighted averaging schemes. In parallel with the aggregator, the Contribution Evaluator assesses the updates using techniques such as simple validation or complex Shapley Value(SV) computation. The Incentive Calculator translates these quality assessments into monetary or reputation-based rewards, often implementing fairness mechanisms. Finally, the Incentive Distributor manages the actual transfer of incentives using economic and game-theoretic techniques such as auction theory, contract theory, or game theory.  The server aims to maximize the utility of both itself and the clients to achieve efficient client selection, efficient reward distribution, efficient resource allocation, and optimized training. As the central authority, the server also ensures fairness throughout the process, enabling every participant to maximize their incentives.

\textbf{Client-Side Architecture:} Each participating client maintains an integrated architecture focused on local computation. Apart from computing local gradients on their data, participants are eager to maximize their individual rewards from the server by optimizing their own strategies. This may happen either with a complete or an incomplete information environment. In contrast, under incomplete information, clients lack knowledge about others’ data quality, effort, or costs and must act under uncertainty. In both cases, they are considered rational, seeking to maximize their payoff by balancing costs and rewards.

The underlying workflow for a C-IBFL may follow principles that will be discussed in the next sections, such as auction-based, contract-based, or game-theory-based. Client selection, contribution evaluation, and incentive distribution will take place based on the underlying principles. In an auction-theory-based incentive framework, clients who bid optimally for a task published by the server are selected for the learning process, serving as an example of the prior explanation.

\subsection{IBFL Architecture for Decentralized Federated Learning (D-IBFL)}

\begin{figure}[ht] 
    \centering
    \includegraphics[width=0.8\textwidth]{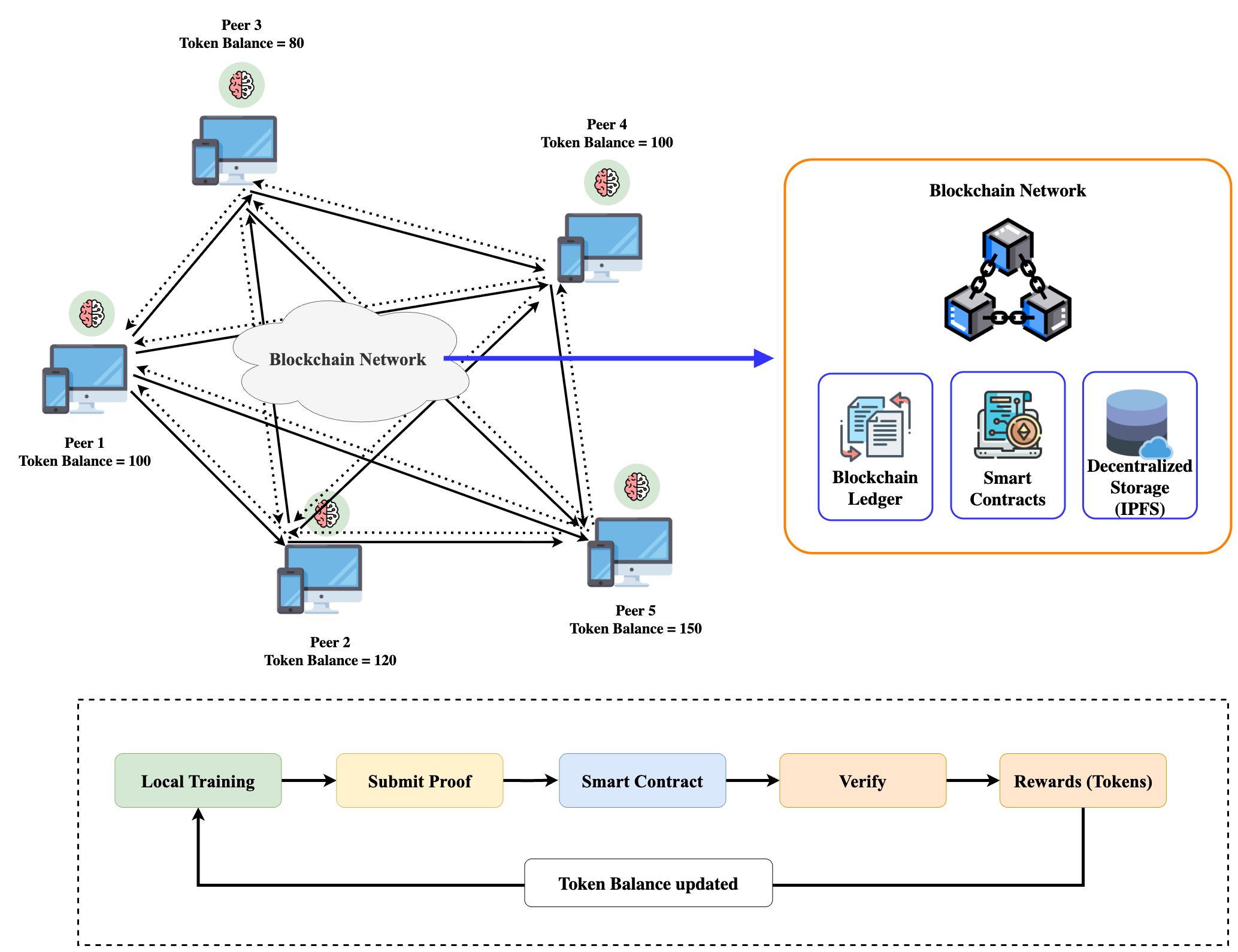} 
    \caption{D-IBFL architecture. Each peer performs local training and exchanges updates in a peer-to-peer manner. A blockchain network, integrated with smart contracts, manages the submission and verification of updates, evaluates contributions, and ensures transparent incentive distribution through token-based rewards.}
    \label{fig:DIBFL}
\end{figure}

Decentralized Federated Learning with Incentives (D-IBFL) eliminates the central coordinator, distributing trust and control across peer nodes. This decentralized architecture addresses the single point of failure (SPoF) vulnerability and the scalability limitations of the client-server architecture. However, it introduces complexity in employing incentive mechanisms, as there is no trusted central authority to evaluate contributions and allocate rewards. The most commonly used techniques in D-IBFL are Blockchain, which provides a transparent reward mechanism, and reputation systems \cite{systematic_review_IFL, BCenabledFLIM_basicpaper}. Figure \ref{fig:DIBFL} illustrates a decentralized incentive mechanism (D-IBFL) powered by blockchain technology.

\textbf{Peer Node Architecture:} Each participating entity follows a peer-to-peer (P2P) architecture for the federated learning process. Each node functions as both a client (training locally) and a server (validating peers). Each node can contain several modules that are important for the learning and incentive processes. The local training module operates identically to C-IBFL clients. If it integrates blockchain, the blockchain acts as a shared, immutable, distributed ledger across the system.

\textbf{Consensus Layer:} D-IBFL requires distributed agreement to verify the updates from peers. This layer often employs blockchain consensus mechanisms (Proof-of-Work (PoW), Proof-of-Stake (PoS)) or Byzantine-fault-tolerant (BFT) protocols to ensure that malicious nodes cannot arbitrarily manipulate outcomes. Through this process, the nodes collectively determine which updates should be integrated into the shared distributed model. At the same time, the consensus layer supports reward allocation by enabling peers to agree on the valuation of each peer’s contribution.

\textbf{Smart Contract–Based Incentive Module:} The most common digital contract found in blockchain-integrated systems with predefined rules. This module enables trustworthy operations without intervention from third parties, as it is an automated process (for example, \textit{"distribute 100 tokens proportionally among nodes whose updates improved validation accuracy by at least 2\%"}). Smart contracts determine and manage incentives (tokens, credits, reputation points). Smart contracts automatically distribute incentives once the updates are validated. Furthermore, the blockchain also provides an audit trail, allowing any participant to verify historical reward distributions and detect manipulation attempts.

\subsection{Comparative Analysis: C-IBFL vs D-IBFL Architectures}

The choice between centralized and decentralized IBFL architectures involves fundamental trade-offs across multiple dimensions:





\begin{table*}[!ht]
\centering
\begin{threeparttable}
\caption{C-IBFL vs D-IBFL: Comparative Analysis}
\resizebox{\textwidth}{!}{%
\begin{tabular}{@{}>{\centering\arraybackslash}p{0.25\textwidth}>{\centering\arraybackslash}p{0.35\textwidth}>{\centering\arraybackslash}p{0.35\textwidth}@{}}
\toprule
\textbf{Aspect} & \textbf{C-IBFL} & \textbf{D-IBFL}  \\
\midrule
Trust Model & Central authority required & Distributed consensus mechanism \\
Incentive Distribution & Server-controlled payment & P2P or smart contract based \\
Scalability & Limited by server capacity & Higher (peer-to-peer scaling) \\
Contribution Verification & Centralized validation & Multi-party verification \\
Privacy Guarantee & Server knows all participants & Enhanced (no single point of knowledge) \\
Implementation Complexity & Lower (simpler coordination) & Higher (consensus overhead) \\
Operational Cost & Server infrastructure costs & Transaction/gas fees per operation \\
Single Point of Failure & Yes (central server) & No (distributed redundancy) \\
Best Use Cases & Healthcare networks, Enterprise settings & Cross-organizational IoT, Blockchains \\
\bottomrule
\end{tabular}%
}
\end{threeparttable}
\end{table*}

A detailed analysis of the terms used in this section will be covered in the following sections.

\section{Taxonomy}

\begin{figure}[ht] 
    \centering
    \includegraphics[width=1\textwidth]{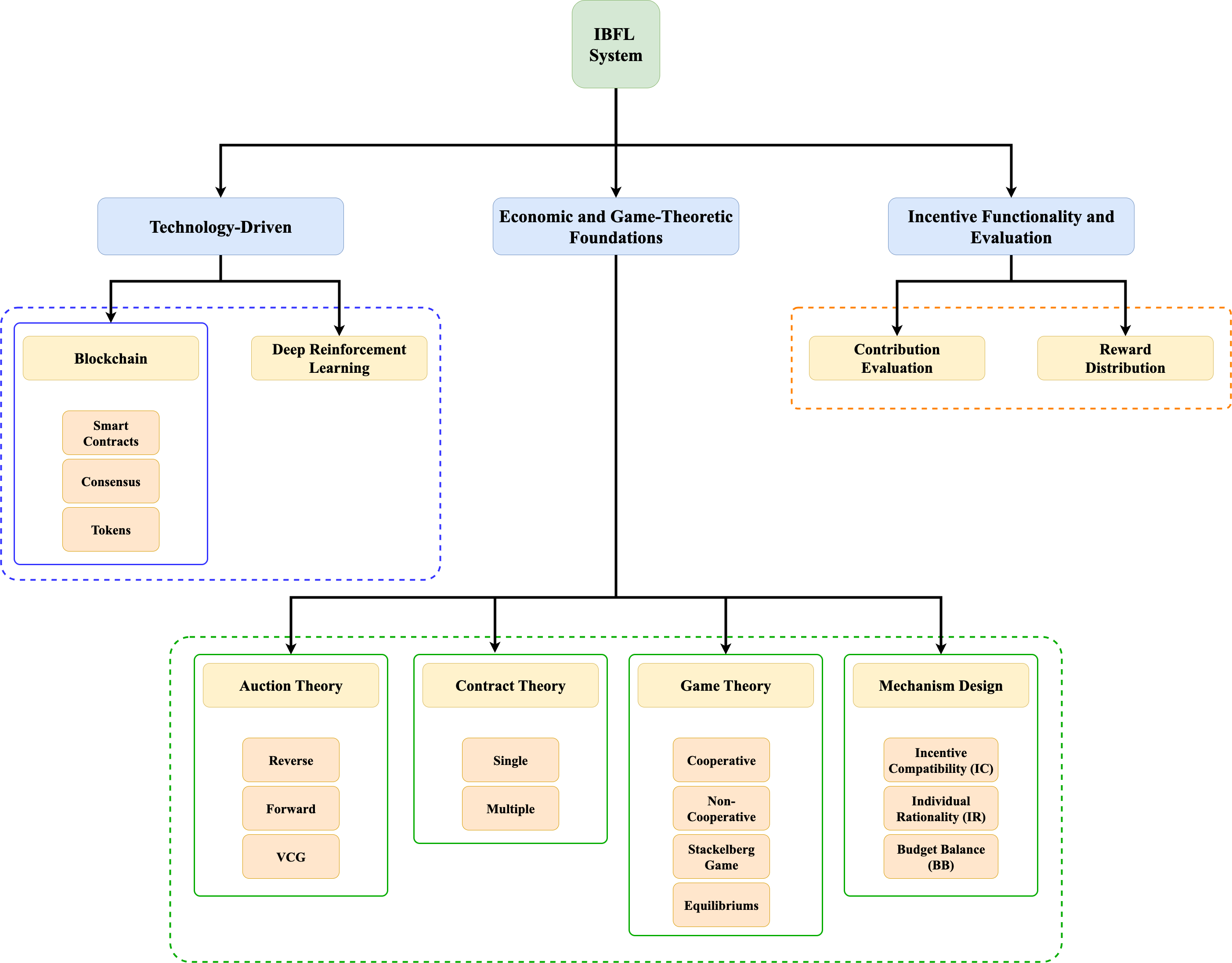} 
    \caption{IBFL taxonomy. Falls under three main categories named Economic and Game-Theoretic Foundations, Technology-Driven approaches, and Functionality}
    \label{fig:taxonomy}
\end{figure}

A taxonomy for Incentive-Based Federated Learning (IBFL) provides a systematic framework to categorize the diverse mechanisms used to encourage participation, ensure fairness, and achieve sustainable collaboration in federated learning environments. Since FL involves self-interested and resource-constrained participating entities, the underlying principles, theoretical foundations, implementation strategies, and evaluation mechanisms are considered in defining an effective taxonomy for incentive mechanisms in a federated learning ecosystem. Figure \ref{fig:taxonomy} illustrates the overview of the taxonomy that will be discussed in this chapter.

\subsection{Economic and Game-Theoretic Foundations}

\subsubsection{Auction Theory}

The theory originates from auctions in economics, where bidders (who offer to pay a certain amount) attempt to acquire or provide commodities (the items being traded by the seller) \cite{economicsmodelsbybuyya}. In the context of federated learning, the most popular architecture for auction-based systems is the client-server architecture. The server (model owner) performs strategy planning for the federated learning system based on the bids, while client devices act as sellers with bidding prices \cite{longtermadaptive_socialwelfaremaximization} in most cases for FL. The commodity being shared in the FL context is local updates. Utility is the difference between the valuation of a certain commodity and its value at the stage of actual payment. The utility is not always a monetary value. It can also include model accuracy, computational cost, and rewards. Every party in the auction is trying to increase its utility. The combination of all participants' utilities can be referred to as collective welfare \cite{longtermadaptive_socialwelfaremaximization}, and the incentive system should always focus on maximizing collective welfare within the federated learning paradigm. The most common auction mechanisms \cite{IntelligentAgentsforAuctionBasedFL_Survey} in federated learning are listed as follows.

\begin{itemize}
    \item \textbf{Forward Auction: } The seller offers the commodity, and several buyers bid for it. The highest bidder wins the auction. The server plays the role of the seller, while clients play the role of the buyer in federated learning. This is less common in the FL paradigm because the server needs clients rather than clients competing for the best server.
    \item \textbf{Reverse Auction: } The buyer announces a demand for the commodity, and sellers compete for it with bids. In the FL context, the server plays the role of the buyer, seeking local updates from clients who act as sellers, with bids corresponding to their costs for computing a given local update. Clients with lower bids for local updates (commodities) win the auction, which is the most common type of auction in federated learning \cite{uncertaintyAwareFLReverseAuction}.
    \item \textbf{Vickrey-Clarke-Groves (VCG) Mechanism: } The VCG mechanism is a type of truthful auction designed to maximize collective welfare \cite{longtermadaptive_socialwelfaremaximization}. Each participant’s payment is determined not by their declared bid but by the externality imposed on others. This is a truth-telling dominant strategy in which participants acquire nothing by lying about their valuation. The VCG introduces fair incentivization for the FL ecosystem while maximizing collective welfare \cite{PVCG_general-purpose_IFL}.
\end{itemize}

\subsubsection{Contract Theory}

 From the aforementioned theories of economics, contract theory studies the design of contracts to motivate participating entities, especially in the presence of asymmetric information. The contract described is essentially an agreement between an employer and an employee in the corporate world. Information asymmetry creates a controversial environment in federated learning, where one party may have greater access to information than the others. A client-server federated learning system can be considered an environment with asymmetric information, where the server does not explicitly know the clients, their computational capabilities, quality levels, data distribution, and so on. A digital contract signed between two parties helps to resolve the information asymmetry mentioned above, where the contract encourages participating entities to perform effectively in return for rewards or reputation scores.

 Bilateral contracts are one-to-one agreements signed between an employee and an employer. Multilateral contracts represent a one-to-many relationship between the employer and multiple employees, which is more suitable for the FL paradigm. The contracts signed can be categorized based on the dimensionality of the characteristics chosen for the contract: one-dimensional contracts, where only one characteristic is considered, and multi-dimensional contracts, where more than one characteristic is considered. In the context of FL, characteristics can include data quality, computational resources, and data volume across clients. The contract can be either static (one-shot), which ends with acceptance or rejection, or repeated (long-term), in which both parties may need to accept or reject the contract again in the future with the necessary amendments \cite{systematic_review_IFL}.

 The contract theory used in centralized federated learning can be identified as the most common approach, as the server acts as an employer and the clients act as employees. Clients who own data tend to choose an optimum contract published by the server. However, designed contracts are more focused on maximizing server-side benefits rather than maximizing collective welfare \cite{IFL_Survey}.

\subsubsection{Game Theory}

Game theory provides a mathematical framework for incentivizing the federated learning paradigm, treating the FL system as a gaming environment \cite{IFL_Survey}. It models participants as rational agents or players, each with strategies and payoffs. A decision made by a player influences both the system and the other players. A few game-specific keywords can be presented as follows;

\begin{itemize}
    \item \textbf{Player:} an individual decision-maker in the game. Each player has a set of possible actions to maximize their own payoff
    \item \textbf{Payoff:} a reward (or outcome) received by a player based on the strategies followed to perform the task. The payoff can be either positive or negative, and it is not only based on the player's own decision but also on the strategies of other players. In FL, the payoff could represent a monetary reward, a contribution to model improvement, or a gain in reputation.
    \item \textbf{Methodology:} the approach or procedure used to model, analyze, and solve a game. This includes defining players, strategies, payoffs, and equilibrium concepts.
    \item \textbf{Equilibrium:} a stable state of the game where no player can benefit from unilaterally changing their strategy. It represents a situation where all players’ strategies are mutually consistent. The most common equilibria in game theory are the Nash equilibrium and the Stackelberg equilibrium, which will be discussed later in this section.
\end{itemize}

\paragraph{Cooperative Games} 

In cooperative game theory, participating entities form coalitions or groups to maximize joint utility rather than individual payoffs. This strategy is common in a federated learning context, where multiple data owners contribute to a better global model. The most common challenge in cooperative games is fairly distributing rewards among coalitions. The Shapley Value, a concept from cooperative game theory, is often used for fair distribution. It considers the average contribution of each participant across all possible coalitions. Furthermore, a cooperative game is a non-zero-sum game in which all participants can benefit simultaneously. In contrast, a player's gain in the game comes at another player's loss, which can be identified as a zero-sum game.

\paragraph{Non-Cooperative Games}

Unlike forming a coalition, participants in non-cooperative games are selfish and always try to maximize their own payoffs through competitive transactions. The approach aligns well with the FL context, where rational players are involved in learning. It can be classified into two categories based on the degree of information exposure. Complete Information, where all players (participants) know all the information about other players, such as strategies, payoffs, and costs. Incomplete Information, where the lack of full access to others' information makes the FL system more realistic, also makes implementing a proper incentive mechanism more challenging.

\paragraph{Nash Equilibrium(NE)} 

\begin{figure}[h] 
    \centering
    \includegraphics[width=0.75\textwidth]{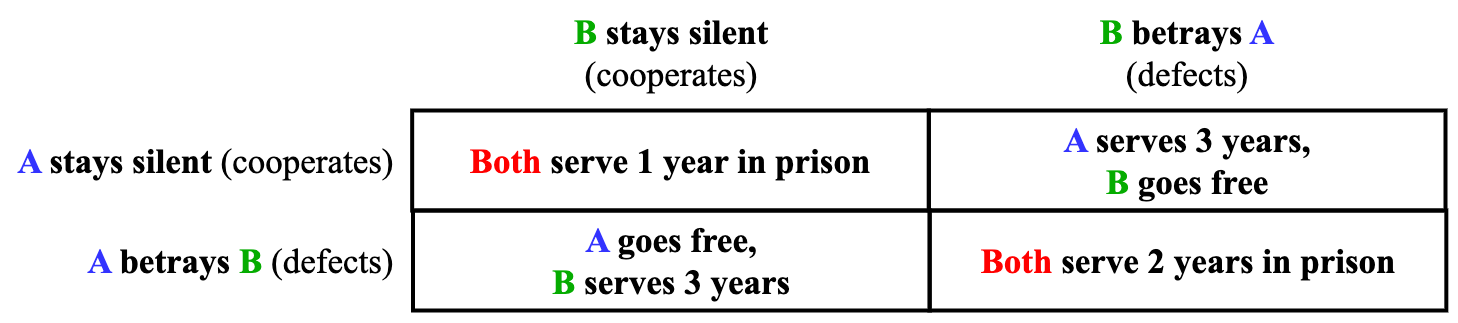} 
    \caption{The \textbf{prisoner dilemma} is a famous game theory concept that illustrates why two rational individuals will not cooperate, even if it seems like the optimal thing for them to do. In the classic scenario, two prisoners are interrogated apart and can confess to the crime (defect) or remain silent (cooperate). Each player's best individual strategy is to defect, but when both defect, they are both worse off than if they had cooperated.}
    \label{fig:prisoner dilemma}
\end{figure}

A Nash equilibrium occurs when no participant can improve their payoff by unilaterally changing their strategy, assuming the strategies of others remain fixed. Nash equilibrium provides a concept of stability in strategic interactions. Figure \ref{fig:prisoner dilemma} represents an example of Nash Equilibrium, where a stable option for both prisoners is to confess to the crime and serve 2 years in prison. In this state, no prisoner can improve their payoff by changing only their own strategy, which defines it as an equilibrium state \cite{SGandDRL_IM_completeandincompleteinfo}.

\paragraph{Stackelberg Game}

\begin{figure}[h] 
    \centering
    \includegraphics[width=0.7\textwidth]{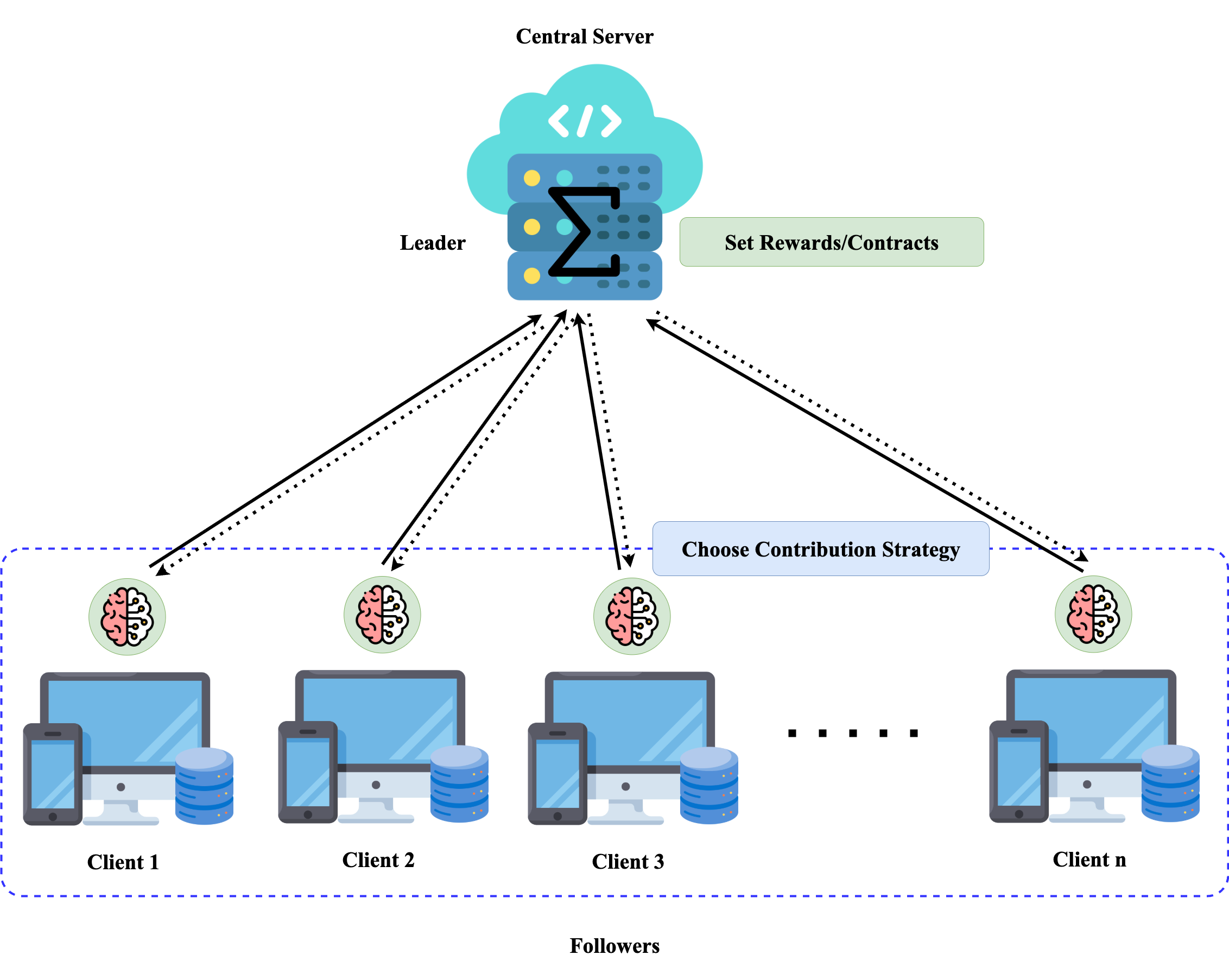} 
    \caption{Repeated Stackelberg game, where the leader chooses an optimal strategy first, and the follower subsequently responds with their best strategy}
    \label{fig:stackelberg_game}
\end{figure}

The Stackelberg Game in the FL context is a hierarchical game with leader and follower roles. In this setting, the leader takes the initial action, after which the followers determine their optimal responses accordingly. In FL, the server plays the role of the leader, and the clients play the roles of followers. In FL, the task publisher often acts as the leader, deciding on reward structures or contracts. Clients (followers) then determine their contribution levels based on these rewards \cite{SGandDRL_IM_completeandincompleteinfo}. The Stackelberg Game allows the leader to anticipate participants' behavior, resulting in more efficient incentive design. Figure \ref{fig:stackelberg_game} illustrates a repeated Stackelberg Game for better understanding.

\paragraph{Stackelberg Equilibrium (SE)} 

The Stackelberg equilibrium occurs when the leader optimally sets the reward and the followers respond in a way that no participant benefits from unilateral deviation. This equilibrium is useful in centralized FL systems, as it ensures efficiency and stability while discouraging free-riding.

\subsubsection{Mechanism Design}

Mechanism design is often referred to as reverse game theory. In game theory, the process typically begins with players, strategies, and payoffs, from which the equilibrium outcome is predicted. In mechanism design, the process begins with the desired outcome (e.g., fairness, participation, efficiency in FL) and then establishes rules (the mechanism) so that rational, self-interested players naturally choose strategies that lead to that outcome \cite{systematic_review_IFL}. In the context of FL, mechanism design ensures that participants truthfully share information while remaining motivated for participation. The key properties of mechanism design \cite{IntelligentAgentsforAuctionBasedFL_Survey} are listed below:

\begin{itemize}
    \item \textbf{Incentive Compatibility (IC): } The best strategy available to participating entities (or players) is to be truthful to the system. Participants should avoid misleading the system by providing incorrect information. These types of mechanisms are called incentive-compatible systems.
    \item \textbf{Individual Rationality (IR): } Ensure that every participant receives at least non-negative utility by participating in the system, which has an incentive mechanism.
    \item \textbf{Budget Balance (BB): } As the term implies, the system does not run a deficit, meaning that the total payments made to participants are less than or equal to the system's income. From the FL perspective (CFL), the server (task publisher) should not spend more on incentives for clients than it has earned.
    \item \textbf{Social Welfare Maximization: } Aiming to maximize the collective welfare of the system.    
\end{itemize}

\subsection{Technology-Driven Mechanisms}

The next section of the taxonomy focuses on technology-driven solutions. Unlike economic or game-theoretic approaches that primarily depend on theoretical models, these approaches leverage emerging technologies to design, deploy, and enforce incentive mechanisms. Two of the most prominent technologies in this category are blockchain and deep reinforcement learning (DRL).

\subsubsection{Blockchain-based Incentive Mechanisms}

Blockchain has become a natural fit for IBFL due to its inherent properties of decentralization, immutability, and transparency. The decentralized nature of blockchain eliminates the issue of a single point of failure introduced by a central server and brings transparency to the system \cite{PoShapley_BCFL}. With its in-built properties, blockchain helps federated learning systems integrate incentive mechanisms, as shown in Figure \ref{fig:DIBFL}. A blockchain network is shared among the participants of the federation and is used to record updates, validate them, manage the model state (or references to it), and distribute rewards. All participants receive a replica of the updated data (blockchain ledger) on the blockchain, which ensures transparency. Blockchain does not need a third party to incentivize participants. This is handled via built-in concepts, which are described below.

\paragraph{Smart Contracts}

A digital contract that allows incentive-related transactions to be conducted securely without the involvement of a trusted intermediary. The reward for a valid local update is calculated, the rewards belonging to the participants are determined, and the agreed-upon rewards are released automatically. An example version of the smart contract is presented in the sample code below. Blockchain helps to record incentives on-chain, ensuring fairness and auditability \cite{BC_basedFramework_ForScalableandIncentivizedFL_Oshani}.

\begin{Verbatim}[fontsize=\footnotesize]
// SPDX-License-Identifier: MIT
pragma solidity ^0.8.0;

contract FLIncentive {
    uint public rewardPerUpdate = 1 ether;
    struct Participant { 
        uint contributions; 
        uint totalReward; 
    }
    mapping(address => Participant) public participants;
    mapping(bytes32 => bool) public submittedUpdates; // Track submitted update hashes

    // Submit a model updated with simple validation
    function submitUpdate(bytes32 updateHash) public {
        // Validation: Reject for duplicate updates
        require(!submittedUpdates[updateHash], "Update already submitted");
        submittedUpdates[updateHash] = true; // Mark as submitted
        participants[msg.sender].contributions += 1;
        participants[msg.sender].totalReward += rewardPerUpdate;
    }

    // Withdraw accumulated rewards - Incentive distribution
    function withdrawReward() public {
        uint amount = participants[msg.sender].totalReward;
        require(amount > 0, "No rewards to withdraw");
        participants[msg.sender].totalReward = 0;
        payable(msg.sender).transfer(amount);
    }
}
\end{Verbatim}

\paragraph{Consensus Mechanisms}

Consensus algorithms ensure that all participants agree on the integrity of model updates and incentive transactions. The most common consensus algorithms are Proof of Work (PoW) and Proof of Stake (PoS). In IBFL, consensus helps prevent malicious behavior, such as fake contributions or duplicate reporting. However, the computational overhead of consensus must be carefully considered, especially in resource-constrained FL environments.

\paragraph{Tokens and Cryptoeconomics}

Tokens (digital currencies or credits) are widely used as transferable incentive units. They not only motivate participants to contribute but can also serve as a medium of exchange between FL participants, service providers, and even external stakeholders. Token-based economies in FL can be further extended to include staking, reputation systems, or marketplace models, allowing for flexible and scalable incentive designs.

\subsubsection{Deep Reinforcement Learning-based (DRL) Incentive Mechanisms}

Deep reinforcement learning (DRL) provides a flexible approach to adjust incentives in real time. Unlike static rules, DRL agents learn policies that determine how to reward participants, select contributors, and adopt incentives under changing conditions. This approach works well in environments where information is incomplete and traditional RL struggles with large state or action spaces \cite{longtermadaptive_socialwelfaremaximization, SGandDRL_IM_completeandincompleteinfo}. DRL uses deep neural networks to represent these policies.

\begin{itemize}
    \item \textbf{Adaptive Incentives:} Continuously balance trade-offs such as accuracy vs. cost or fairness vs. efficiency by learning from participant behavior.
    \item \textbf{Robustness Against Strategic Players:} Dynamically adjust rewards, making it harder for participants to exploit the system.
    \item \textbf{Scalability:} Handle large numbers of participants and heterogeneous environments where fixed mechanisms fail.
\end{itemize}

In summary, blockchain ensures trust and transparency, while DRL provides flexibility and adaptability, making it a strong combination for incentive-based FL.

\subsection{Incentive Functionality and Evaluation}

Evaluating contributions and distributing rewards are critical, as even the most well-designed incentive architecture will fail without a fair evaluation of contributions and effective reward allocation.

\paragraph{Contribution Evaluation} Measure the value each participant adds to the global model (CFL) or the consensus/personalized model (DFL). Since federated learning involves heterogeneous data and devices, naive methods such as counting updates are insufficient. Common approaches include the Shapley Value, a game-theoretic method to fairly allocate rewards based on each participant’s marginal contribution \cite{PoShapley_BCFL}; model accuracy contribution, which evaluates a participant’s impact by measuring the difference in accuracy when their updates are included versus excluded; resource-aware contribution, which considers computation, communication, and energy costs alongside accuracy improvements; and reputation-based contribution, which assigns a higher weight to participants with historically consistent and high-quality contributions. A robust evaluation system must be fair, secure against manipulation, and scalable to ensure participant trust and system sustainability.

\paragraph{Reward Distribution} Once contributions are evaluated, rewards must be distributed fairly, in a motivating manner, and while balancing the budget \cite{BC_basedFramework_ForScalableandIncentivizedFL_Oshani}. Distribution strategies vary depending on the FL context. Proportional distribution allocates rewards based on contribution scores, such as accuracy gain or data quality, while threshold-based distribution provides rewards only if contributions exceed a predefined level, discouraging free-riding. Hybrid approaches combine proportional and fixed components, ensuring both fairness and guaranteed minimum rewards to satisfy individual rationality. Dynamic distribution adapts incentives over time based on participant consistency, network conditions, or model performance requirements. All reward strategies should follow core principles, such as Incentive Compatibility, Individual Rationality, and Budget Balance.


\section{Application-Oriented Incentive Mechanisms in Federated Learning}

It is not enough to limit ourselves to the theoretical foundations discussed in previous sections because federated learning schemes can be applied across various application domains, from critical healthcare systems to technology-driven distributed systems with proper incentivization. This section examines how incentive mechanisms are adapted to address the specific requirements, constraints, and opportunities present in four key application domains: healthcare, smart cities and IoT ecosystems, vehicular networks, and blockchain-based decentralized systems.

\subsection{Healthcare and Medical AI:}

The healthcare domain can be considered a critical application for a federated learning paradigm because the data utilized by the FL system should be handled carefully due to privacy concerns and regulations related to patient privacy protection. Another important challenge in healthcare is sustainable participation across the federated learning system. A proper incentive mechanism can resolve these challenges, and the ways existing approaches provide solutions to them are listed below.

\paragraph{Class-Specific Contribution Evaluation}
Tastan et al. \cite{Tastan} proposed ShapFed, a Class-Specific Shapley Value (CSSV)  for a contribution evaluation strategy that helps address class imbalance issues in medical datasets. Traditional frameworks for federated learning consider the full dataset completely in computing accuracy in contribution evaluation, which is not effective in most cases because certain disease categories may be underrepresented and remain hidden when considering the full setup. Furthermore, they determined the alignment of the model from the last layer to reduce computational complexity. Apart from that, they introduced ShapFed-WA, a novel weighted average algorithm that serves as a replacement for FedAvg \cite{FedAvg}, leveraging the normalized contribution factor from CSSV. That method outperformed on class-imbalanced medical image datasets (Chest X-Ray \cite{ChestXRay} and Fed-ISIC2019 \cite{Fed-isic2019}), demonstrating significant improvements in both utility and fairness, which makes it particularly suitable for medical imaging applications where data distribution heterogeneity is common.

\paragraph{Emergency Response and UAV-Assisted Medical Infrastructure}
Li et al. \cite{UAV_FL_SGIM} combines hierarchical federated learning with Unmanned Aerial Vehicle (UAV)-assisted mobile edge computing, creating a dynamic system capable of responding to emergencies while protecting patient privacy. Their approach addressed the response delay issue in the federated learning process by adapting the UAV layer as a local aggregator to overcome the sudden traffic surges caused by massive data generation and emergency responses near hospitals. The incentive mechanism is modeled as a two-layer Stackelberg game involving three participants: user devices (patients), UAVs (mobile edge servers), and base stations.  This game-theoretic approach determines optimal data requirements, participant selection, and reward allocation to maximize the utility of each participant while ensuring high-quality model training and privacy. The framework explicitly addresses the unwillingness of users and UAVs to consume computational and communication resources without satisfactory rewards, providing dynamic incentive structures that adapt to emergency scenarios. A combination of a hierarchical incentive mechanism-empowered federated learning with UAV-assisted mobile edge computing enables a sustainable healthcare system, even in critical situations, while maintaining data privacy.

\subsection{Internet of Things and Smart Infrastructure}

With the growth of smart devices and sensors, smart infrastructures powered by the IoT are emerging and being utilized in many federated learning paradigms. However, the heterogeneous and dynamic data generated by this infrastructure must be carefully handled with a proper incentive mechanism to maintain sustainable participation across the system.

\paragraph{General-Purpose VCG-Based Mechanisms}
Cong et al.\cite{PVCG_general-purpose_IFL} proposed a procurement-VCG-based (PVCG) incentive mechanism that follows the fundamentals of a reverse auction. Furthermore, this application considers capacity limits for suppliers, typically data owners. This application works well in the area of general-purpose federated learning systems, including smart city applications. Moreover, they interpolated unseen data, a common practice in most federated learning systems, particularly in smart cities, where devices and sensors often do not share accurate data or share data at all. The PVCG mechanism ensures incentive compatibility, individual rationality, and weak budget balance, making it particularly suitable for multi-stakeholder IoT ecosystems where participants have diverse objectives. Ensuring truthful reporting of participants’ costs and capabilities, this application works well in heterogeneous environments such as smart infrastructures.

\paragraph{Uncertainty-Aware Auctions}
Xu et al. \cite{uncertaintyAwareFLReverseAuction} addressed a critical yet often overlooked challenge of uncertainty in IoT-based federated learning. Their proposed solution employed a reverse-auction mechanism, optimizing expected social welfare to tackle system and client uncertainties. However, the application is built on assumptions, such as a static environment and the notion that the aggregation interval for a round is sufficient for all clients to complete local training. Selected participants from the ecosystem received payment at the end of each learning round, ensuring sustainable and motivated participation in areas such as smart environments, where clients frequently face resource constraints and intermittent connectivity. Incorporating Incentive Compatibility and Individual Rationality jointly, this VCG-based application ensured truthfulness across the paradigm.

\paragraph{Blockchain-Enabled IoT Data Sharing}
Cai et al. \cite{BCbasedFLforIoTSharing} proposed a comprehensive framework for IoT data sharing that combines federated learning with a consortium blockchain, a model-sharing platform, IoT users, and edge nodes. They leveraged two blockchain networks for distinct purposes: tracking transactions and maintaining records of reputation. The reputation mechanism punishes users who fail to complete model-sharing tasks, ensuring data quality and reliable participation. The deep reinforcement learning integration for model selection enabled the maximization of long-term social welfare across the proposed framework. This approach is particularly relevant for IoT applications spanning industrial systems, smart grids, and intelligent transportation, where data quality directly impacts system performance and safety. The underlying blockchain infrastructure and learning-based DRL optimization allow this framework to serve as a sustainable solution across smart industrial systems.

\subsection{Vehicular Networks and Edge Computing}

Vehicular networks present distinct challenges when considered as a federated learning system, including high mobility, intermittent connectivity, dynamic client populations, real-time processing requirements, and the need for long-term support. Therefore, the incentive mechanisms proposed for this domain must be adapted to the dynamic environment and supported over the long term.

\paragraph{Multi-Branch Strategy Network}
Wu et al. \cite{longtermadaptive_socialwelfaremaximization} addressed the challenge of client shifting in vehicular networks, where vehicles regularly join and leave the training process due to mobility. As a solution to the lack of support for long-term optimization, their approach incorporated VCG auctions with long-term availability, where bids were determined based on the costs of communication, computation, and data usage. They introduced deep reinforcement learning as an experience-driven solution to the problem of complete information unavailability in federated learning. However, the above DRL-based approach is affected by the client shifting problem. A multi-branch strategy network, composed of a common feature extractor and a specialized decision head, is presented, allowing the system to handle client shifting while maintaining sustainable incentivized federated learning and optimizing resources. Altogether, the proposed approach satisfies both incentive compatibility and individual rationality simultaneously, aiming to maximize social welfare, especially in the case of client shifting within environments of incomplete information, such as highly dynamic vehicular networks over the long term.

\paragraph{Edge-Cloud Collaborative Resource Allocation}
Liu et al. \cite{IoV_auctionbased} focused on edge-cloud collaboration for vehicular networks, proposing a truthful reverse auction mechanism for utility maximization. This work primarily focuses on computational and bandwidth constraints in vehicular networks, where vehicles have heterogeneous resources and data quality, while edge servers have limited capacity. The mechanism optimally allocates training tasks between edge and cloud resources, considering network latency, computational capabilities, and data utility. By formulating the problem as a utility maximization, the reverse auction framework provides an incentive mechanism for vehicles to truthfully report their capabilities and costs, thereby enabling efficient resource utilization across the edge–cloud spectrum.

\paragraph{Incentivized Data Dissemination in Vehicular Edge Computing}
Bute et al. \cite{IMforIoV} examined incentive-based federated learning data dissemination specifically for vehicular edge computing networks. A reverse auction mechanism, including bid determination and winner selection, which maximizes social welfare, is leveraged as a solution to the recognized unwillingness of vehicles to share data and computational resources without proper compensation. The greedy algorithm-based approach provides computational efficiency that is suitable for real-time vehicular applications, where decision-making must occur within strict latency constraints. This framework addresses the unique challenge of data dissemination in high-mobility environments where participants have limited interaction time with edge infrastructure.

\subsection{Blockchain-Based Decentralized Systems}

Blockchain technology enables trusted transactions without the need for a single trusted authority, providing a transparent, decentralized infrastructure with inherent features such as traceability and immutability. These built-in features, consensus protocols, and blockchain economics are particularly valuable when selecting an appropriate incentive mechanism for federated learning, especially within a decentralized architecture.

\paragraph{Proof of Shapley Value for Contribution-Based Consensus}
Cheng et al. \cite{PoShapley_BCFL} introduced PoShapley-BCFL, a combination of blockchain and the Shapley value that enables an incentivized federated learning paradigm for decentralized architectures. This framework ensures fairness, truthfulness, and transparency through inherited features from both the Shapley Value and blockchain. PoShapley-BCFL leveraged the Proof-of-Shapley Value consensus mechanism to evaluate contributions from participants using Monte Carlo sampling, making the mechanism lightweight, unlike the classic Shapley Value. Furthermore, the consensus approach allowed for the selection of an aggregator for each round. These contribution values are further used to formulate a robust aggregation weight, which is encrypted using RSA to defend against low-quality data attacks, thereby addressing both contribution fairness and model robustness simultaneously. By enabling contribution evaluation in a decentralized manner through blockchain integration, PoShapley-BCFL becomes suitable for scalable decentralized federated learning systems, such as decentralized marketplaces or decentralized banking, which require transparent transactions and fairness throughout the process while preserving privacy.

\paragraph{Smart Contract-Based Automation and Hybrid Incentives}
Wu and Seneviratne \cite{BC_basedFramework_ForScalableandIncentivizedFL_Oshani} presented a blockchain-enabled approach that addresses the challenge of encouraging meaningful contributions in decentralized environments, especially for training large language models. The integrated smart contracts are responsible for automating client registration, updating validation, reward distribution, and maintaining the transparency of the global state. Their approach consists of three components: an on-chain alignment for real-time reward generation, an off-chain fairness check to offload computational complexity from the blockchain, and a consistency multiplier for long-term, high-quality, and sustainable engagement. The gas cost analysis shows that the framework is economically feasible across different scales, offering useful guidance for real-world blockchain-based federated learning systems.

\paragraph{Blockchain-Enabled Data Sharing Incentives}
Wang et al. \cite{BCenabledFLIM_basicpaper} proposed a blockchain-empowered framework that integrates a dual incentive scheme (reputation-based and SV-based) for decentralized and multi-party data sharing scenarios. The blockchain network they proposed consists of a federated learning requester, model participants, a computing center, an aggregation node, and smart contracts. The reputation system enables punishments for malicious nodes, making the framework more secure, while SV-based incentives ensure fairness. They outsourced the Shapley Value computation to the computing center, as it incurs computational complexity on the blockchain. This approach is particularly relevant for scenarios involving multiple organizations with competing interests but mutual benefits from collaborative learning.\\

\begin{table*}[!htbp]
\centering
\caption{Comparison of Application-Oriented Federated Learning Incentive Mechanisms. }
\label{tab:fl_comparison}
\begin{tabular}{@{}>{\centering\arraybackslash}p{0.15\textwidth}>{\centering\arraybackslash}p{0.18\textwidth}>{\centering\arraybackslash}p{0.25\textwidth}>{\centering\arraybackslash}p{0.37\textwidth}@{}}
\toprule
\textbf{Work} & \textbf{Application Domain} & \textbf{Incentive Mechanism Category} & \textbf{Incentive Concept / Core Idea} \\
\midrule

Tastan et al. \cite{Tastan} & Healthcare and Medical AI & Game Theory (Shapley Value) & Contribution evaluation, fairness, robust aggregation \\

Li et al. \cite{UAV_FL_SGIM} & Healthcare and Medical AI  & Stackelberg Game & Hierarchical reward allocation for UAV–MEC healthcare \\

Cong et al. \cite{PVCG_general-purpose_IFL} & Smart Infrastructure & VCG Auction & Fair bidding and welfare-maximizing participation \\

Xu et al. \cite{uncertaintyAwareFLReverseAuction} & Smart Infrastructure & Reverse Auction + VCG & Truthful client selection under training uncertainty \\

Cai et al. \cite{BCbasedFLforIoTSharing} & Smart Infrastructure & Blockchain + DRL & Online reputation-based IoT data sharing incentive \\

Zhan et al. \cite{SGandDRL_IM_completeandincompleteinfo} & General IoT & DRL + Game Theory (Stackelberg Game) & Adaptive reward optimization under incomplete information  \\

Wu et al. \cite{longtermadaptive_socialwelfaremaximization} & Vehicular Networks & VCG Auction + DRL & Long-term adaptive incentive for periodic client shifts \\

Liu et al. \cite{IoV_auctionbased} & Vehicular Networks & Reverse Auction & Utility-maximizing edge–cloud collaboration incentive \\

Bute et al. \cite{IMforIoV} & Vehicular Networks & Reverse Auction & Data dissemination incentive using greedy optimization \\

Cheng et al. \cite{PoShapley_BCFL} & Blockchain-Based Systems & Blockchain + Shapley Value & Proof-of-Shapley consensus for fair contribution  \\

Wang et al. \cite{BCenabledFLIM_basicpaper} & Blockchain-Based Systems & Blockchain + Shapley Value & Transparent reward distribution for data sharing with reputation analysis \\

Wu and Seneviratne \cite{BC_basedFramework_ForScalableandIncentivizedFL_Oshani} & Blockchain-Based Systems & Blockchain + Shapley Value & Hybrid on/off-chain rewards, transparency, alignment-based incentive  \\

\bottomrule
\end{tabular}
\end{table*}

Table \ref{tab:fl_comparison} provides a comparative overview of application-oriented incentive mechanisms in federated learning, highlighting the paper, target application domain, incentive mechanism category, and the core concept or approach used to encourage participation and ensure fairness across diverse settings such as healthcare, IoT, vehicular networks, and blockchain-based systems.


\section{Challenges and Future Directions}

Even though much of the existing literature attempts to address challenges in federated learning from the perspective of incentive mechanisms, several potential challenges still remain unexplored or underexplored. 

\subsection{Challenges}

The most common challenge encountered is the limitation of scalable and privacy-preserving contribution evaluation. The classic Shapley Value computation for fair incentives is computationally expensive. Still, many studies have focused on reducing the complexity of federated learning by either offloading it as a separate task or approximating it. However, balancing accuracy, cost, and privacy (through secure aggregation and verifiable computation) remains an open challenge. 

As discussed in the previous section, most existing approaches consider either a static participation setting (VCG, Stackelberg Game) or a dynamic one with unrealistic assumptions, such as clients not leaving the learning round until it finishes, but being allowed to leave only after completing a round. These assumptions and static settings can be considered theoretical solutions rather than practical approaches to real-world problems. DRL-based frameworks attempt to address the dynamic nature, but they still incur significant computational complexity that needs to be reduced. A dynamic federated learning setting with participant churn and nonstationary costs may pose challenges when developing an incentive mechanism, and there is still plenty of room to address these challenges.

Most prominent solutions are either centralized or hierarchical in nature, which may over-optimize utility for the central authority rather than for the overall system. Creating a mechanism that does not violate budget constraints while ensuring truthfulness and fairness could be a critical challenge. Furthermore, due to the single point of failure inherent in centralized architectures, most applications are now adapting to decentralized architectures, which eliminate reliance on a single authority. Blockchain offers a prominent solution to this problem, leveraging its inherent immutability, traceability, and transferability. However, on-chain costs, computational complexity, and latency create barriers. Even with hybrid on/off-chain designs, which reduce overhead, there are still complicated correctness proofs.

Federated learning itself is a research domain with multiple subfields, each of which can be formulated as distinct research questions. Integrating an incentive mechanism should not only optimize itself but also jointly optimize other subfields. Sticking to single-factor optimization while making assumptions about the others poses an open challenge in achieving holistic and practical solutions in federated learning systems.
Furthermore, robustness to strategic behaviors and malicious updates remains a critical concern. While incentives can motivate participation, they may also encourage strategic behavior, such as poisoning or Sybil attacks. Addressing this requires a careful integration of robust aggregation methods, reputation systems, and incentive design.

\subsection{Future Directions}

Several promising research directions remain open for incentive mechanisms in federated learning. One of them is the development of provably efficient Shapley value estimates with privacy guarantees to address the computational complexity issue caused by exact SV. Future research needs to focus on designing estimators with allocation error bounds under secure aggregation or differential privacy while considering the trade-offs of the system. This is achievable using a mix of techniques in Shapley estimation and secure multiparty computation.

Another important direction is establishing economic-theoretic guarantees for learning-based controllers. Architectures integrating DRL adaptation with strict economic constraints, such as approximate incentive-compatibility and long-term budget balance, need to be developed. The long-term adaptive VCG+DRL approach is a good starting point for this.

Hybrid blockchain designs also provide a strong research trajectory. Off-chain processing paired with on-chain settlement, potentially supported by zero-knowledge proofs for correctness, can reduce gas costs without compromising privacy and accountability in reward systems.

Mechanisms resilient to non-IID data and heterogeneous contribution metrics are crucial, as practical federated learning often involves skewed data distributions. Incentive metrics should be based on model utility rather than merely computational contribution, and evaluations must follow standardized benchmarks and protocols.

Finally, the solution for security, reputation, and economic mechanisms is left as an open challenge. Such joint designs must be capable of punishing or countering malicious updates while maintaining participation incentives. Such mechanisms can leverage a combination of reputation mechanisms, cryptographic attestations, and tailored payments to ensure fairness, stability, and ongoing engagement.


\section{Summary and Conclusions}

Making federated learning work in the real world requires more than just clever algorithms. It requires an understanding of the rational behavior of participating entities and economic incentives. This chapter has shown that incentive design lies at the core of the practical viability of federated learning. Without proper incentives, participating entities may simply refuse to contribute or free-ride on the system.

Moreover, auction mechanisms and VCG approaches promote truthful behavior, while Shapley values provide a mathematically fair method for reward distribution. Contract and game theory facilitate the management of strategic interactions among self-interested participants. Blockchain ensures transparency and automatic settlement without a central authority by leveraging its inherent features, while deep reinforcement learning adds flexibility to handle dynamic and uncertain environments as an experience-driven solution, performing effectively even under incomplete information. Each approach comes with its benefits and trade-offs, and the choice largely depends on the specific application context.

Real-world applications in healthcare, smart infrastructure, vehicular networks, and blockchain-based systems demonstrate that these approaches go beyond basic proof of concept. They address practical challenges, ranging from protecting patient privacy during medical AI training to incentivizing IoT devices in smart infrastructure, including smart cities, smart transportation, and smart grids. Nevertheless, significant challenges remain. Computing fair contributions at scale while preserving privacy is computationally expensive. Handling the dynamic participation under realistic assumptions about client behavior requires further research. Balancing the advantages of blockchain-enabled systems with their costs and latency remains a critical challenge in this research domain. Most importantly, there is a need for end-to-end solutions that simultaneously optimize incentives alongside other federated learning objectives, such as robustness, privacy, and efficiency.

The future directions depend on scalable contribution evaluation mechanisms, hybrid blockchain protocols that reduce on-chain latency, and learning-based controllers with provable economic guarantees. Only then can federated learning systems be truly ready for mass deployment, systems in which participation is not only technically feasible but also economically viable and genuinely sustainable.


\bibliographystyle{IEEEtran}
\bibliography{bcfl_references}

\begin{thebibliography}{10}
\providecommand{\url}[1]{#1}
\csname url@samestyle\endcsname
\providecommand{\newblock}{\relax}
\providecommand{\bibinfo}[2]{#2}
\providecommand{\BIBentrySTDinterwordspacing}{\spaceskip=0pt\relax}
\providecommand{\BIBentryALTinterwordstretchfactor}{4}
\providecommand{\BIBentryALTinterwordspacing}{\spaceskip=\fontdimen2\font plus
\BIBentryALTinterwordstretchfactor\fontdimen3\font minus \fontdimen4\font\relax}
\providecommand{\BIBforeignlanguage}[2]{{%
\expandafter\ifx\csname l@#1\endcsname\relax
\typeout{** WARNING: IEEEtran.bst: No hyphenation pattern has been}%
\typeout{** loaded for the language `#1'. Using the pattern for}%
\typeout{** the default language instead.}%
\else
\language=\csname l@#1\endcsname
\fi
#2}}
\providecommand{\BIBdecl}{\relax}
\BIBdecl

\bibitem{FL-challenges-and-future-direction}
T.~Li, A.~K. Sahu, A.~Talwalkar, and V.~Smith, ``Federated learning: Challenges, methods, and future directions,'' \emph{IEEE Signal Processing Magazine}, vol.~37, no.~3, pp. 50--60, 2020.

\bibitem{advances-and-open-problems-in-fl}
P.~Kairouz, H.~McMahan, B.~Avent, A.~Bellet, M.~Bennis, A.~Bhagoji, K.~Bonawitz, Z.~Charles, G.~Cormode, R.~Cummings, R.~D'Oliveira, S.~El~Rouayheb, D.~Evans, J.~Gardner, Z.~Garrett, A.~Gascón, B.~Ghazi, P.~Gibbons, M.~Gruteser, and S.~Zhao, ``Advances and open problems in federated learning,'' 12 2019.

\bibitem{survey-IM}
X.~Tu, K.~Zhu, N.~C. Luong, D.~Niyato, Y.~Zhang, and J.~Li, ``Incentive mechanisms for federated learning: From economic and game theoretic perspective,'' \emph{IEEE Transactions on Cognitive Communications and Networking}, vol.~8, no.~3, pp. 1566--1593, 2022.

\bibitem{pmlr-v54-mcmahan17a}
B.~McMahan, E.~Moore, D.~Ramage, S.~Hampson, and B.~A.~y. Arcas, ``{Communication-Efficient Learning of Deep Networks from Decentralized Data},'' in \emph{Proceedings of the 20th International Conference on Artificial Intelligence and Statistics}, ser. Proceedings of Machine Learning Research, A.~Singh and J.~Zhu, Eds., vol.~54.\hskip 1em plus 0.5em minus 0.4em\relax PMLR, 20--22 Apr 2017, pp. 1273--1282.

\bibitem{IM-for-FL}
X.~Tu, K.~Zhu, C.~Nguyen, D.~Niyato, Y.~Zhang, and J.~Li, ``Incentive mechanisms for federated learning: From economic and game theoretic perspective,'' \emph{IEEE Transactions on Cognitive Communications and Networking}, vol.~8, pp. 1566--1593, 09 2022.

\bibitem{pmlr-v119-sim20a}
R.~H.~L. Sim, Y.~Zhang, M.~C. Chan, and B.~K.~H. Low, ``Collaborative machine learning with incentive-aware model rewards,'' in \emph{Proceedings of the 37th International Conference on Machine Learning}, ser. Proceedings of Machine Learning Research, H.~D. III and A.~Singh, Eds., vol. 119.\hskip 1em plus 0.5em minus 0.4em\relax PMLR, 13--18 Jul 2020, pp. 8927--8936.

\bibitem{game-theory-based-BC}
Y.~He, M.~Luo, B.~Wu, L.~Sun, Y.~Wu, Z.~Liu, and K.~Xiao, ``A game theory-based incentive mechanism for collaborative security of federated learning in energy blockchain environment,'' \emph{IEEE Internet of Things Journal}, vol.~10, no.~24, pp. 21\,294--21\,308, 2023.

\bibitem{pmlr-v130-fraboni21a}
Y.~Fraboni, R.~Vidal, and M.~Lorenzi, ``Free-rider attacks on model aggregation in federated learning,'' in \emph{Proceedings of The 24th International Conference on Artificial Intelligence and Statistics}, ser. Proceedings of Machine Learning Research, A.~Banerjee and K.~Fukumizu, Eds., vol. 130.\hskip 1em plus 0.5em minus 0.4em\relax PMLR, 13--15 Apr 2021, pp. 1846--1854.

\bibitem{lin2019freeridersfederatedlearningattacks}
\BIBentryALTinterwordspacing
J.~Lin, M.~Du, and J.~Liu, ``Free-riders in federated learning: Attacks and defenses,'' 2019. [Online]. Available: \url{https://arxiv.org/abs/1911.12560}
\BIBentrySTDinterwordspacing

\bibitem{overcomingfree-ridingbehaviorinpeer-to-peersystems}
M.~Feldman and J.~Chuang, ``Overcoming free-riding behavior in peer-to-peer systems,'' \emph{SIGecom Exch.}, vol.~5, no.~4, p. 41–50, Jul. 2005.

\bibitem{economicsmodelsbybuyya}
R.~Buyya, D.~Abramson, J.~Giddy, and H.~Stockinger, ``Economic models for resource management and scheduling in grid computing,'' \emph{Concurrency and Computation: Practice and Experience}, vol.~14, no. 13-15, pp. 1507--1542, 2002.

\bibitem{grideconomy_buyya}
R.~Buyya, D.~Abramson, and S.~Venugopal, ``The grid economy,'' \emph{Proceedings of the IEEE}, vol.~93, no.~3, pp. 698--714, 2005.

\bibitem{buyya2002economicbaseddistributedresourcemanagement}
\BIBentryALTinterwordspacing
R.~Buyya, ``Economic-based distributed resource management and scheduling for grid computing,'' \emph{Monash University, Melbourne, Australia}, 2002. [Online]. Available: \url{https://arxiv.org/abs/cs/0204048}
\BIBentrySTDinterwordspacing

\bibitem{FedAvg}
B.~McMahan, E.~Moore, D.~Ramage, S.~Hampson, and B.~A.~y. Arcas, ``{Communication-Efficient Learning of Deep Networks from Decentralized Data},'' in \emph{Proceedings of the 20th International Conference on Artificial Intelligence and Statistics}, ser. Proceedings of Machine Learning Research, A.~Singh and J.~Zhu, Eds., vol.~54.\hskip 1em plus 0.5em minus 0.4em\relax PMLR, 20--22 Apr 2017, pp. 1273--1282.

\bibitem{systematic_review_IFL}
A.~Ali, I.~Ilahi, A.~Qayyum, I.~Mohammed, A.~Al-Fuqaha, and J.~Qadir, ``A systematic review of federated learning incentive mechanisms and associated security challenges,'' \emph{Comput. Sci. Rev.}, vol.~50, no.~C, Nov. 2023.

\bibitem{BCenabledFLIM_basicpaper}
Z.~Wang, B.~Yan, and A.~Dong, ``Blockchain empowered federated learning for data sharing incentive mechanism,'' \emph{Procedia Computer Science}, vol. 202, pp. 348--353, 05 2022.

\bibitem{longtermadaptive_socialwelfaremaximization}
L.~Wu, S.~Guo, Z.~Hong, Y.~Liu, W.~Xu, and Y.~Zhan, ``Long-term adaptive vcg auction mechanism for sustainable federated learning with periodical client shifting,'' \emph{IEEE Transactions on Mobile Computing}, vol.~23, no.~5, pp. 6060--6073, 2024.

\bibitem{IntelligentAgentsforAuctionBasedFL_Survey}
X.~Tang, H.~Yu, X.~Li, and S.~Kraus, ``Intelligent agents for auction-based federated learning: a survey,'' in \emph{Proceedings of the Thirty-Third International Joint Conference on Artificial Intelligence}, ser. IJCAI '24, 2024.

\bibitem{uncertaintyAwareFLReverseAuction}
J.~Xu, B.~Tang, H.~Cui, and B.~Ye, ``An uncertainty-aware auction mechanism for federated learning,'' in \emph{Algorithms and Architectures for Parallel Processing}, Z.~Tari, K.~Li, and H.~Wu, Eds.\hskip 1em plus 0.5em minus 0.4em\relax Singapore: Springer Nature Singapore, 2024, pp. 1--18.

\bibitem{PVCG_general-purpose_IFL}
\BIBentryALTinterwordspacing
M.~Cong, H.~Yu, X.~Weng, J.~Qu, Y.~Liu, and S.~Yiu, ``A vcg-based fair incentive mechanism for federated learning,'' \emph{CoRR}, vol. abs/2008.06680, 2020. [Online]. Available: \url{https://arxiv.org/abs/2008.06680}
\BIBentrySTDinterwordspacing

\bibitem{IFL_Survey}
A.~K. Nair, S.~Coleri, J.~Sahoo, L.~R. Cenkeramaddi, and E.~D. Raj, ``Incentivized federated learning: A survey,'' \emph{IEEE Transactions on Emerging Topics in Computational Intelligence}, vol.~9, no.~5, pp. 3190--3209, 2025.

\bibitem{SGandDRL_IM_completeandincompleteinfo}
Y.~Zhan, P.~Li, Z.~Qu, D.~Zeng, and S.~Guo, ``A learning-based incentive mechanism for federated learning,'' \emph{IEEE Internet of Things Journal}, vol.~7, no.~7, pp. 6360--6368, 2020.

\bibitem{PoShapley_BCFL}
Z.~Cheng, Y.~Liu, C.~Wu, Y.~Pan, L.~Zhao, and C.~Zhu, ``Poshapley-bcfl: A fair and robust decentralized federated learning based on blockchain and the proof of shapley-value,'' in \emph{Neural Information Processing: 30th International Conference, ICONIP 2023, Changsha, China, November 20–23, 2023, Proceedings, Part II}.\hskip 1em plus 0.5em minus 0.4em\relax Berlin, Heidelberg: Springer-Verlag, 2023, p. 531–549.

\bibitem{BC_basedFramework_ForScalableandIncentivizedFL_Oshani}
B.~Wu and O.~Seneviratne, ``Blockchain-based framework for scalable and incentivized federated learning,'' in \emph{Companion Proceedings of the ACM on Web Conference 2025}, ser. WWW '25.\hskip 1em plus 0.5em minus 0.4em\relax New York, NY, USA: Association for Computing Machinery, 2025, p. 1761–1767.

\bibitem{Tastan}
N.~Tastan, S.~Fares, T.~Aremu, S.~Horvath, and K.~Nandakumar, ``Redefining contributions: shapley-driven federated learning,'' in \emph{Proceedings of the Thirty-Third International Joint Conference on Artificial Intelligence}, ser. IJCAI '24, 2024.

\bibitem{ChestXRay}
T.~Rahman, A.~Khandakar, M.~Kadir, K.~Islam, K.~Islam, R.~Mazhar, T.~Hamid, M.~Islam, S.~Kashem, Z.~Mahbub, M.~Ayari, and M.~Chowdhury, ``\BIBforeignlanguage{English}{Reliable tuberculosis detection using chest x-ray with deep learning, segmentation and visualization},'' \emph{\BIBforeignlanguage{English}{IEEE Access}}, vol.~8, pp. 191\,586--191\,601, Oct. 2020.

\bibitem{Fed-isic2019}
J.~O. du~Terrail, S.-S. Ayed, E.~Cyffers, F.~Grimberg, C.~He, R.~Loeb, P.~Mangold, T.~Marchand, O.~Marfoq, E.~Mushtaq, B.~Muzellec, C.~Philippenko, S.~Silva, M.~Tele\'{n}czuk, S.~Albarqouni, S.~Avestimehr, A.~Bellet, A.~Dieuleveut, M.~Jaggi, S.~P. Karimireddy, M.~Lorenzi, G.~Neglia, M.~Tommasi, and M.~Andreux, ``Flamby: datasets and benchmarks for cross-silo federated learning in realistic healthcare settings,'' in \emph{Proceedings of the 36th International Conference on Neural Information Processing Systems}, ser. NIPS '22.\hskip 1em plus 0.5em minus 0.4em\relax Red Hook, NY, USA: Curran Associates Inc., 2022.

\bibitem{UAV_FL_SGIM}
C.~Li, M.~Song, and Y.~Luo, ``Federated learning based on stackelberg game in unmanned-aerial-vehicle-enabled mobile edge computing,'' \emph{Expert Systems with Applications}, vol. 235, p. 121023, 2024.

\bibitem{BCbasedFLforIoTSharing}
T.~Cai, X.~Li, W.~Chen, Z.~Wei, and Z.~Ye, ``Blockchain-based federated learning for iot sharing: Incentive scheme with reputation mechanism,'' in \emph{Blockchain and Trustworthy Systems}, J.~Chen, B.~Wen, and T.~Chen, Eds.\hskip 1em plus 0.5em minus 0.4em\relax Singapore: Springer Nature Singapore, 2024, pp. 270--284.

\bibitem{IoV_auctionbased}
L.~Liu, J.~Zhang, Z.~Wang, and J.~Xu, ``A truthful reverse auction mechanism for federated learning utility maximization resource allocation in edge–cloud collaboration,'' \emph{Mathematics}, vol.~11, p. 4968, 12 2023.

\bibitem{IMforIoV}
M.~S. Bute, P.~Fan, and Q.~Luo, ``Incentive based federated learning data dissemination for vehicular edge computing networks,'' in \emph{Proceedings of the 98th IEEE Vehicular Technology Conference (VTC 2023)}, 2023, pp. 1--5.

\end{thebibliography}

\end{document}